\documentclass[letterpaper,journal]{IEEEtran}
\usepackage{amsmath,amsfonts}
\usepackage{algorithmic}
\usepackage{array}
\usepackage[caption=false,font=normalsize,labelfont=sf,textfont=sf]{subfig}
\usepackage{textcomp}
\usepackage{stfloats}
\usepackage{url}
\usepackage{verbatim}
\usepackage[final]{graphicx}
\usepackage{capt-of}
\usepackage{comment}
\usepackage{bm}
\usepackage{cite}
\usepackage{booktabs}
\usepackage{multirow}
\usepackage{makecell}
\usepackage{float}
\usepackage{threeparttable}
\usepackage[
    colorlinks=true,
    linkcolor=black,
    citecolor=black,
    urlcolor=blue
]{hyperref}
\usepackage{capt-of}

\begin{document}

\title{Learning Geometry-Aware Planning Representations for Monocular Flight with Cross-Modal Supervision}

\author{Wencan Feng, Bailing Tian, Yulin Hui, Zhiyu Li, and Junjie Lu$^{*}$
\thanks{All authors are with the School of Electrical and Information Engineering, Tianjin University, Tianjin, 300072, China.}
\thanks{The source code: \url{https://github.com/TJU-Aerial-Robotics/GeoMonoPlan}}}

\IEEEaftertitletext{
    \vspace{-3mm}
    \vspace{-0.5em}
    \begin{minipage}{\textwidth}
        \centering
        \includegraphics[width=\textwidth]{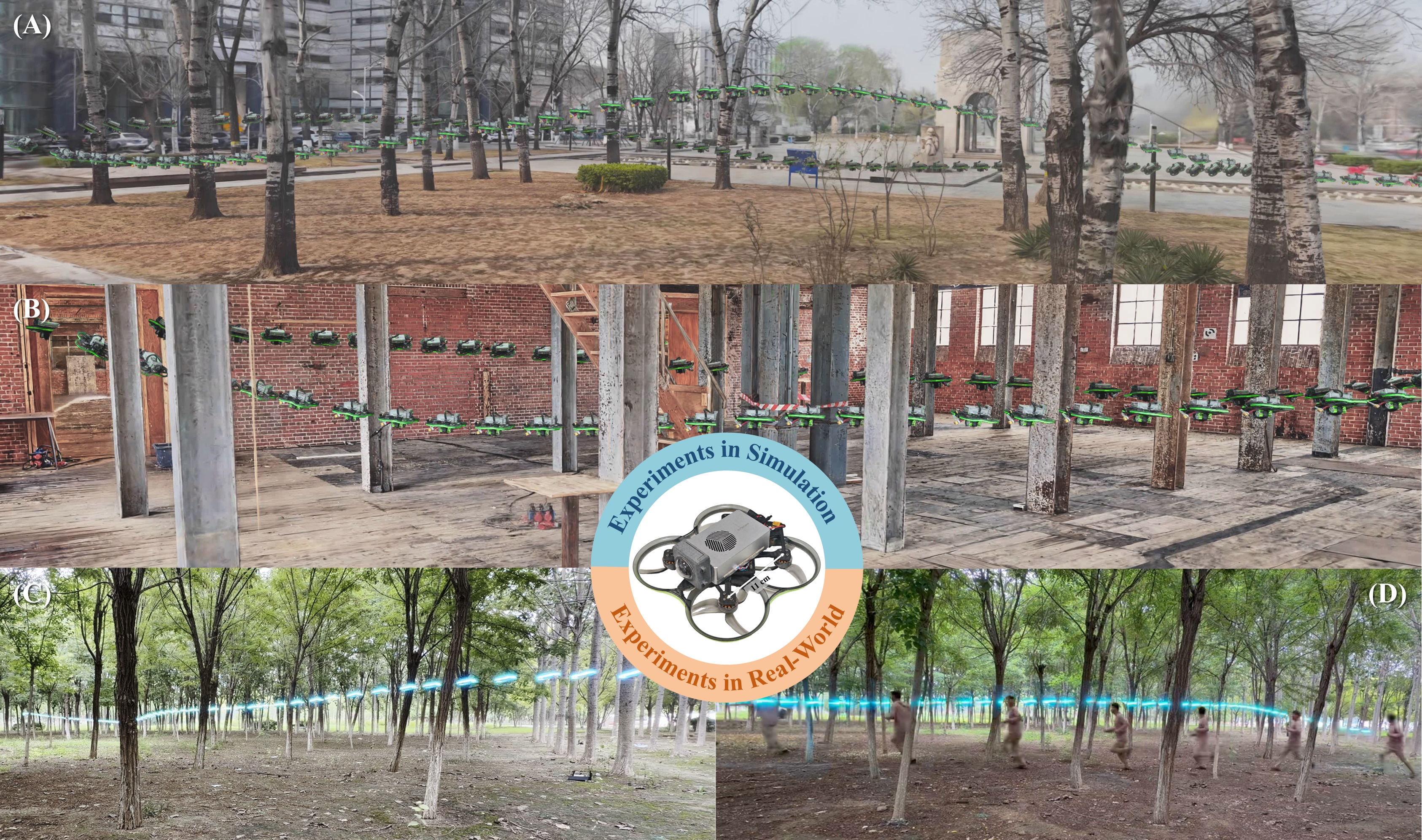}
        \vspace{-5mm}
        \captionof{figure}{Overview of the proposed method. (A) Simulation flight in the forest. (B) Simulation flight in the industrial facility environment. (C) Real-world obstacle-avoidance flight in a forest environment. (D) Real-world tracking flight.}
        \label{fig:overview}
    \end{minipage}
    \vspace{0.5em}
}

\maketitle

\begin{abstract}
Existing high-speed navigation systems for micro aerial vehicles (MAVs) commonly rely on depth cameras, stereo vision, or LiDAR to obtain explicit geometric information, while monocular cameras offer a lighter and lower-cost alternative. However, monocular flight remains challenging because RGB images offer no direct metric cues and visual appearance varies considerably across environments. We propose a geometry-aware monocular visual planning framework that predicts trajectories from two image frames, their relative pose, and the MAV state. Without explicitly reconstructing depth, the proposed method complements the downstream planning objective with feature-level geometric supervision by transferring the planning-relevant representation learned by a privileged depth expert to the RGB planner through cross-modal distillation. The framework first trains a depth-based geometric expert, then distills its planning features into a monocular branch using paired RGB-D observations, and finally adapts the planner to the downstream trajectory planning objective. This design encourages the policy to learn planning-relevant geometric features, enabling collision avoidance. We further use a 3D Gaussian Splatting (3DGS) simulator to render visual observations for data collection, paired RGB-D supervision, and simulation evaluation. The planner trained in simulation is then deployed in real-world environments without real-world fine-tuning, achieving obstacle avoidance at speeds of up to 6.5 m/s.
\end{abstract}


\section{Introduction}
Achieving agile autonomous flight in highly cluttered environments remains a long-standing challenge for MAVs. Most existing autonomous navigation systems rely on sensors that provide explicit metric geometry, such as LiDARs and depth cameras, to perceive the surrounding environment, followed by well-established mapping-and-planning pipelines or end-to-end policy networks to generate collision-aware trajectories \cite{r4,r8,r9}. While these approaches have demonstrated impressive performance, their reliance on dedicated geometric sensors constrains the compactness and design flexibility of miniature aerial platforms. In contrast, monocular RGB planning requires only a lightweight commodity camera, offering greater flexibility under stringent payload and energy budgets.

Biological vision demonstrates that agile flight does not necessarily rely on explicit geometric perception. Small flying animals such as bees and forest-dwelling birds can rapidly navigate through densely cluttered environments using only two-dimensional visual cues, such as object scale changes and scene contours, without explicitly reconstructing the surrounding geometry\cite{r6,r7}. Likewise, experienced first-person-view (FPV) pilots can execute aggressive maneuvers and traverse narrow gaps using only a monocular video stream \cite{r5}. These observations indicate that monocular images contain sufficient information to support agile autonomous flight, making monocular cameras an attractive sensing modality for MAVs. Compared with LiDARs and depth cameras, monocular cameras are lightweight, inexpensive, easy to integrate, and naturally compatible with wide field-of-view optics such as fisheye lenses. Moreover, they provide rich texture and contextual information that is unavailable to purely geometric sensors.

Despite these advantages, learning agile navigation policies directly from these images remains fundamentally challenging. Most existing navigation methods employ imitation learning or reinforcement learning to train RGB policies directly using downstream task objectives \cite{r11,r12}, without explicitly constraining their internal representations to preserve planning-relevant geometry. As a result, they often suffer from appearance overfitting and poor cross-scene generalization. Unlike depth sensors, monocular cameras provide no explicit metric distance information, making it difficult to infer the traversable spatial structure required for motion planning. Furthermore, these images are highly sensitive to illumination, textures, and object appearance, forcing the navigation policy to distinguish geometry relevant to obstacle avoidance from appearance variations that are irrelevant to planning. To alleviate these issues, some approaches introduce explicit geometric intermediate representations, such as depth  \cite{r16,r19} or optical flow \cite{r14,r17}, to extract geometry-aware cues from observations, while others exploit privileged depth supervision or ground-truth geometric information during training to guide policy learning \cite{r15,r18}. However, explicit geometric intermediate representations incur additional computational overhead and inference latency. Directly trained RGB policies, on the other hand, rely primarily on downstream task supervision, without explicitly encouraging the learned representations to preserve planning-relevant geometric information. Consequently, these representations may remain sensitive to appearance variations and may not reliably capture the geometric structure required for collision avoidance.

To address these limitations, we propose a geometry-aware monocular navigation framework that takes two consecutive RGB images, their relative pose, and the MAV state as input. It should be noted that the primary objective of monocular navigation is not to recover pixel-wise accurate depth, but to learn geometry-aware representations that preserve planning-relevant structural information. Guided by this insight, the framework narrows the geometric gap between a privileged depth expert and a monocular planner by transferring planning-oriented geometric representations through feature-level distillation. During training, the depth expert learns geometry-aware planning features under privileged supervision, while the student network reproduces these structured representations via feature alignment. To further encourage the learned representations to preserve planning-relevant geometry and reduce their dependence on appearance cues, we introduce a lightweight coarse-depth prediction branch with an auxiliary training objective. This auxiliary task provides additional geometric regularization for the distilled representations. Furthermore, we develop a scalable simulation pipeline based on 3DGS, which provides photorealistic images, corresponding geometric supervision, and map-based trajectory cost labels for data generation and training. During deployment, the planner operates solely on two consecutive monocular images and MAV states, without requiring any depth estimation module or additional geometric sensors.
The main contributions of this work are summarized as follows:

\begin{itemize}

\item We propose a framework that transfers planning-oriented geometric representations from a privileged depth expert to the student planner, aided by a lightweight coarse-depth branch that reinforces geometric structure. 

\item We develop a monocular planner that plans trajectories from sequential images, relative pose, and MAV states, and performs online evaluation to enable agile, collision-aware flight.

\item We establish a 3DGS-based data pipeline that provides aligned RGB observations, privileged geometry, vehicle states, and trajectory-cost labels for training and evaluation.

\end{itemize}

\section{Related Work}
\subsection{Navigation with Explicit Geometric Perception}

Most autonomous navigation systems rely on depth cameras or LiDAR to obtain explicit geometric information. Classical pipelines construct ESDF maps from depth observations and combine path search with trajectory optimization \cite{r21, r25}, whereas methods based on safe flight corridors (SFCs) optimize trajectories within convex collision-free corridors \cite{r20}. Ren et al. jointly optimized trajectory safety and efficiency using LiDAR, enabling high speed autonomous flight \cite{r9}.

Learning-based planners further reduce latency by directly mapping sensor observations to trajectories or controls. Loquercio et al. used imitation learning to predict planning outputs from depth images and MAV states \cite{r8}. Without expert demonstrations, Lu et al.'s Guidance Learning trains a polynomial-trajectory predictor by back-propagating gradients of the trajectory objective \cite{r2,r33}. Zhang et al. combined differentiable physics with reinforcement learning to train a GRU policy from depth observations and vehicle states for agile multi-MAV flight \cite{r24}. Xu et al. fused LiDAR range images with temporal point-flow features in an RL policy that directly generates obstacle-avoidance controls \cite{r22}.

\subsection{Monocular Vision-based Navigation}

Monocular cameras are lightweight and power-efficient, making them suitable for resource-constrained MAVs. Early methods directly mapped RGB images to navigation commands using CNN \cite{r11, r12}. However, their lack of explicit geometric supervision may limit policy reliability and generalization. Other approaches recover geometric cues as intermediate representations. MonoNav estimates depth and reconstructs local geometry before planning \cite{r16}, whereas Hu et al. use optical flow to represent scene depth and structure for robust and agile navigation \cite{r14}.

Sim-to-real transfer remains another challenge for monocular navigation. CAD2RL combines deep reinforcement learning with domain randomization to transfer policies to real MAVs \cite{r13}. More recently, 3DGS has enabled rapid, high-fidelity reconstruction of real scenes for robotic simulation \cite{r28}. Compared with manually constructed game-engine environments, it provides realistic visual observations with less scene-building effort. Low et al.'s SOUS VIDE uses 3DGS to collect data for end-to-end policies deployed on real MAVs \cite{r29}. Huang et al. further combine 3DGS environments with domain adaptation and randomization to improve real-world generalization \cite{r15}.

\section{Method}

\subsection{Problem Formulation}\label{sec:problem}

We study monocular visual trajectory planning for a quadrotor. At each planning step, the planner maps the visual and state observation to a single smooth polynomial trajectory that drives the robot toward a goal.

The planner receives two types of inputs.
(i) A pair of temporal images $\{I_{k-1},I_k\}$ and their relative camera pose, represented by a unit quaternion $\mathbf{q}\in\mathbb{R}^{4}$ and a metric translation $\mathbf{t}\in\mathbb{R}^{3}$, which together provide geometric cues for scene understanding.
(ii) A body-frame state observation $\mathbf{x}_k=[\mathbf{v}_b,\mathbf{a}_b,\mathbf{g}_b]\in\mathbb{R}^{9}$, comprising the current velocity, acceleration, and goal vector.

Following the one-stage planning paradigm~\cite{r2}, the network simultaneously predicts $N$ candidate motion primitives. For each primitive $i\in\{1,\dots,N\}$ it outputs an end state $\mathcal{S}_f^{(i)}$, a selection score $s^{(i)}$, and a predicted collision cost $\hat{c}^{(i)}$.

We first use $s^{(i)}$ only to shortlist a top-$K$ candidate set, and then select within it the primitive that minimizes a trajectory cost $J^{(i)}$ combining the predicted collision cost $\hat{c}^{(i)}$ with the closed-form dynamics terms evaluated online. The same cost structure $J^{(i)}$ is also used as the training objective, ensuring that online selection follows the same optimality criterion as training.

The $N$ candidate primitives are associated with anchors on a polar lattice spanning the camera field of view. Each primitive is realized by a single-segment quintic polynomial $\mathbf{p}^{(i)}:[0,T]\!\to\!\mathbb{R}^{3}$ that matches the current state $\mathcal{S}_0=[\mathbf{p}_0,\mathbf{v}_0,\mathbf{a}_0]$ and the predicted end state $\mathcal{S}_f^{(i)}=[\mathbf{p}_f^{(i)},\mathbf{v}_f^{(i)},\mathbf{a}_f^{(i)}]$. The polynomial coefficients admit a closed-form solution in $\mathcal{S}_f^{(i)}$ for fixed $\mathcal{S}_0$ and $T$; trajectory costs such as the jerk integrals are therefore differentiable in the end state and are reused both as training objectives and as the closed-form selection cost at inference. Detailed derivations of the polynomial coefficients and the cost gradients with respect to the terminal state are provided in~\cite{r30,r31}.

\begin{figure}[t]
    \centering
    \includegraphics[width=0.8\linewidth]{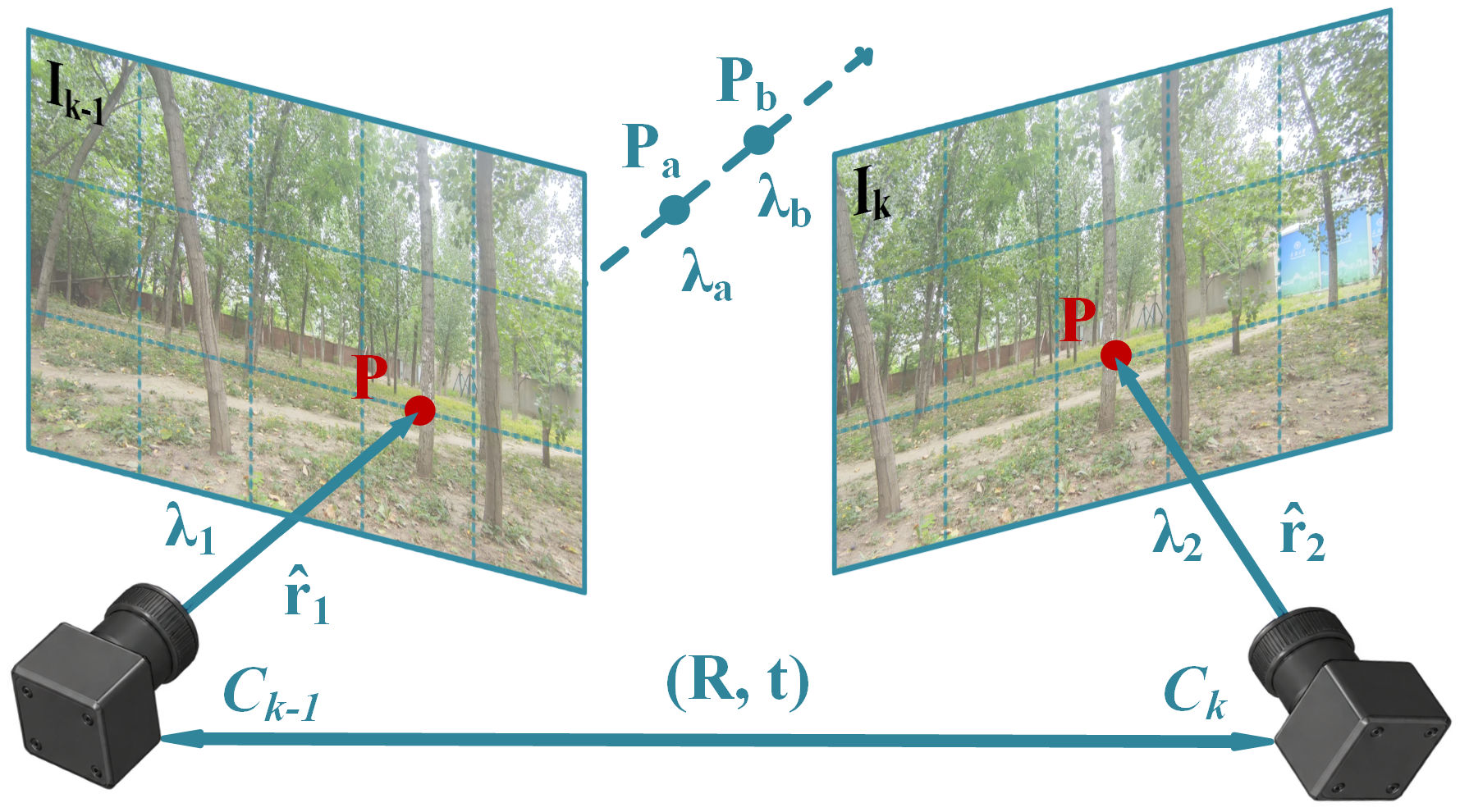}
    \caption{Geometric motivation for the proposed input. A single image observation leaves the metric depth of a scene point ambiguous, as illustrated by the candidate positions $P_a$ and $P_b$ along the same viewing ray. The second observation and the known metric relative pose provide the complementary metric-scale geometric information.}
    \label{fig:pose}
    \vspace{-3mm}
\end{figure}

As shown in Fig.~\ref{fig:pose}, we further note that the proposed input provides complementary metric-scale cues that are unavailable from a single image. This analysis does not imply that dense geometry is always recoverable in practice; rather, it shows that metric relative translation provides a scale cue under non-degenerate motion and visibility conditions. Our method is designed to exploit this cue without explicitly establishing correspondences. Consider a static scene point $\mathbf{P}$ observed by two views with known metric relative pose $(\mathbf{R},\mathbf{t})\in SE(3)$, where $\mathbf{R}=\mathcal{R}(\mathbf{q})\in SO(3)$ is the rotation matrix corresponding to the quaternion $\mathbf{q}$, and let $\hat{\mathbf{r}}_1,\hat{\mathbf{r}}_2$ be its unit viewing rays. With a single view, $\mathbf{P}=\lambda\hat{\mathbf{r}}_1$ is determined only up to a positive scale, since $\mathbf{P}\sim s\mathbf{P}$ projects identically. Therefore, metric depth is unidentifiable from one image alone. With two views, the depths $(\lambda_1,\lambda_2)$ along the rays satisfy equation:
\begin{equation}
\bigg[\,- \mathbf{R}\hat{\mathbf{r}}_1\;\;\hat{\mathbf{r}}_2\,\bigg]
\begin{bmatrix}\lambda_1\\ \lambda_2\end{bmatrix}=\mathbf{t}.
\label{eq:two_view}
\end{equation}
Equation~\eqref{eq:two_view} uniquely determines both $\lambda_1$ and $\lambda_2$ whenever the rays are non-degenerate. By linearity, the solution for $\lambda_1$ can be expressed as:
\begin{equation}
\lambda_1=\|\mathbf{t}\|\,\psi(\hat{\mathbf{r}}_1,\hat{\mathbf{r}}_2,R,\hat{\mathbf{t}}),
\qquad \hat{\mathbf{t}}=\mathbf{t}/\|\mathbf{t}\|.
\end{equation}
So the entire metric scale of the depth is carried by the baseline magnitude $\|\mathbf{t}\|$, while $\psi$ is a dimensionless function of the viewing directions and pose. Hence, two frames with a metric relative pose resolve the monocular scale ambiguity and provide sufficient information to determine metric geometry in principle, under the standard assumptions of a static, textured scene visible in both views.

\subsection{Network Architecture}

\begin{figure*}[t]
    \centering
    \includegraphics[width=\textwidth]{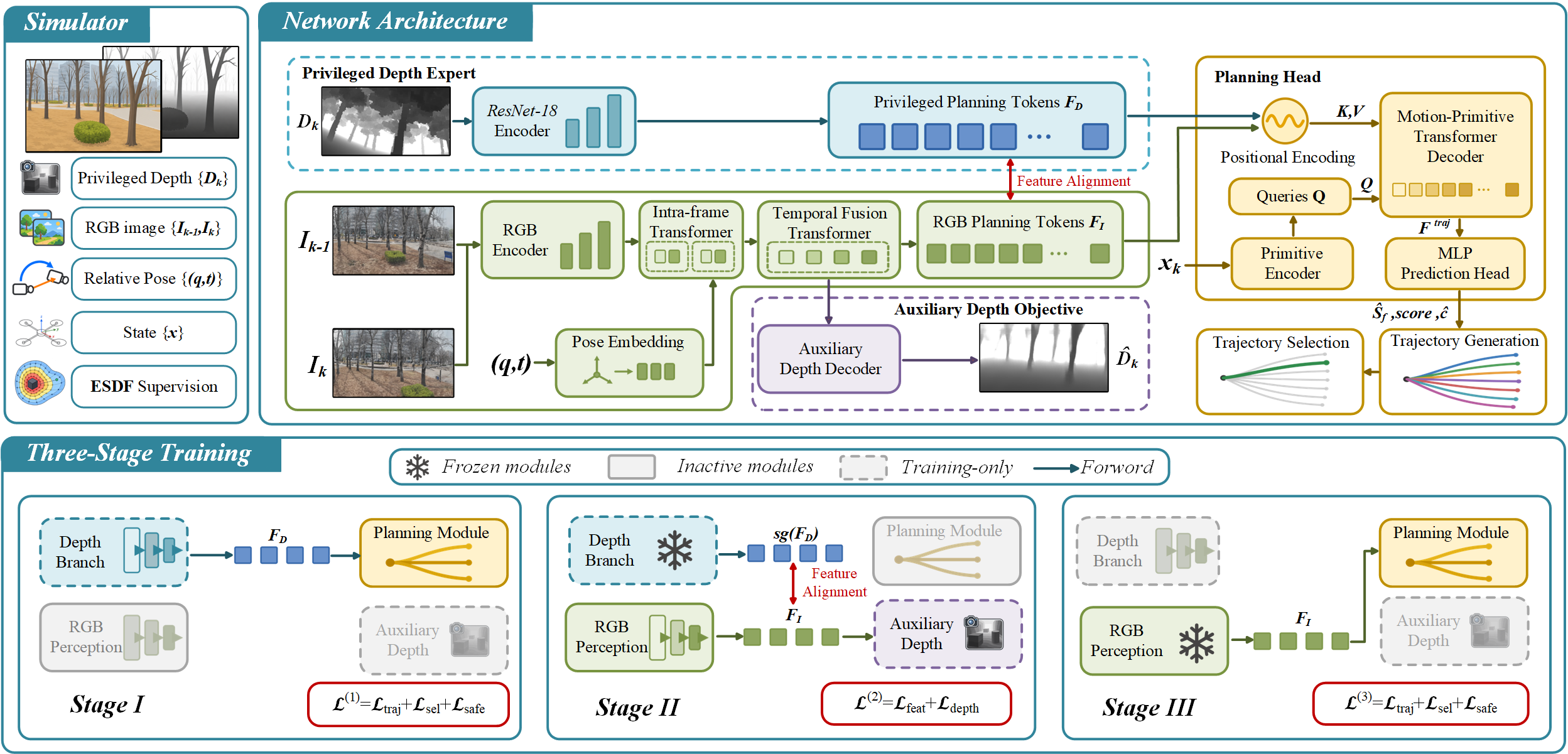}
    \vspace{-6mm}
    \caption{Network architecture and progressive three-stage training strategy of the proposed method. The framework collects simulation data and uses planning-feature alignment, auxiliary depth supervision, and progressive training to train a monocular planner.}
    \vspace{-2mm}
    \label{fig:Network frame}
\end{figure*}

The proposed method consists of a privileged depth expert, a monocular RGB student, and a planning head shared by the two branches. Rather than forcing the RGB branch to reproduce all details of a depth map, we use a compact planning representation as the bridge between the two modalities. Both branches produce planning features $\mathbf{F}\in\mathbb{R}^{L_p\times d}$ with the same spatial layout and feature dimension, where $L_p$ denotes the number of spatial planning tokens and $d$ is the embedding dimension. This shared representation allows both branches to use the same motion-primitive decoder, making the cross-modal alignment directly relevant to downstream planning.

\subsubsection{Privileged depth branch}
Given the current normalized depth observation $D_k$, the depth branch uses a ResNet-18 encoder to extract geometry-aware features. The resulting feature map is mapped to a compact spatial grid and flattened into planning tokens, denoted by $\mathbf{F}_D\in\mathbb{R}^{L_p\times d}$.  This compact representation serves as the privileged target for feature distillation.  Before being passed to the shared motion-primitive decoder, a standard sinusoidal positional encoding is added to preserve the spatial arrangement of the tokens. The depth branch is used only during training and is not required for deployment.

\subsubsection{Pose-conditioned temporal RGB branch}
The student receives two consecutive images and their metric relative pose.  A monocular image encoder extracts the frame-level spatial features $\mathbf{F}_{k-1}^{\mathrm{img}}$, $\mathbf{F}_{k}^{\mathrm{img}}$.  In contrast to the depth encoder, the RGB encoder retains a higher spatial resolution so that thin obstacles and image details are not prematurely discarded. The relative pose is embedded by a learned linear map,
\begin{equation}
    \mathbf{e}_{\mathrm{pose}}
    =\mathbf{W}_{\mathrm{pose}}
    [\mathbf{q}^{\top},\mathbf{t}^{\top}]^{\top}
    +\mathbf{b}_{\mathrm{pose}}
    \in\mathbb{R}^{d}.
    \label{eq:pose_embedding}
\end{equation}
The pose embedding is then broadcast to every token of both frames.  The pose-conditioned tokens are then processed by a shared intra-frame Transformer encoder $\mathcal{T}_{\mathrm{frame}}$:
\begin{equation}
    \mathbf{F}_{j}^{\mathrm{frame}}
    =\mathcal{T}_{\mathrm{frame}}\!\left(
    \operatorname{PE}(
    \mathbf{F}_{j}^{\mathrm{img}}
    +\mathbf{e}_{\mathrm{pose}})
    \right),
\end{equation}
where $\operatorname{PE}(\cdot)$ denotes the standard sinusoidal positional encoding and $j\in\{k-1,k\}$ indexes the two input frames.

After the intra-frame encoder models long-range spatial dependencies within each view, we concatenate the two encoded views along the feature dimension and apply a second Transformer encoder for temporal fusion:
\begin{equation}
    \mathbf{F}_{I}=\mathcal{T}_{\mathrm{fus}}\!\left(
    [\mathbf{F}_{k-1}^{\mathrm{frame}}\,\Vert\,\mathbf{F}_{k}^{\mathrm{frame}}]
    \right).
    \label{eq:temporal_fusion}
\end{equation}

A convolutional adapter maps the fused representation to the same spatial feature layout as the depth branch.  After flattening, the resulting RGB planning tokens are denoted by $\mathbf{F}_I\in\mathbb{R}^{L_p\times d}$. Consequently, relative motion supplies the metric baseline missing from a single image, while attention permits the network to associate geometric cues across the two observations without explicitly constructing a point cloud or a depth map.

\begin{figure*}[t]
    \centering
    \includegraphics[width=\textwidth]{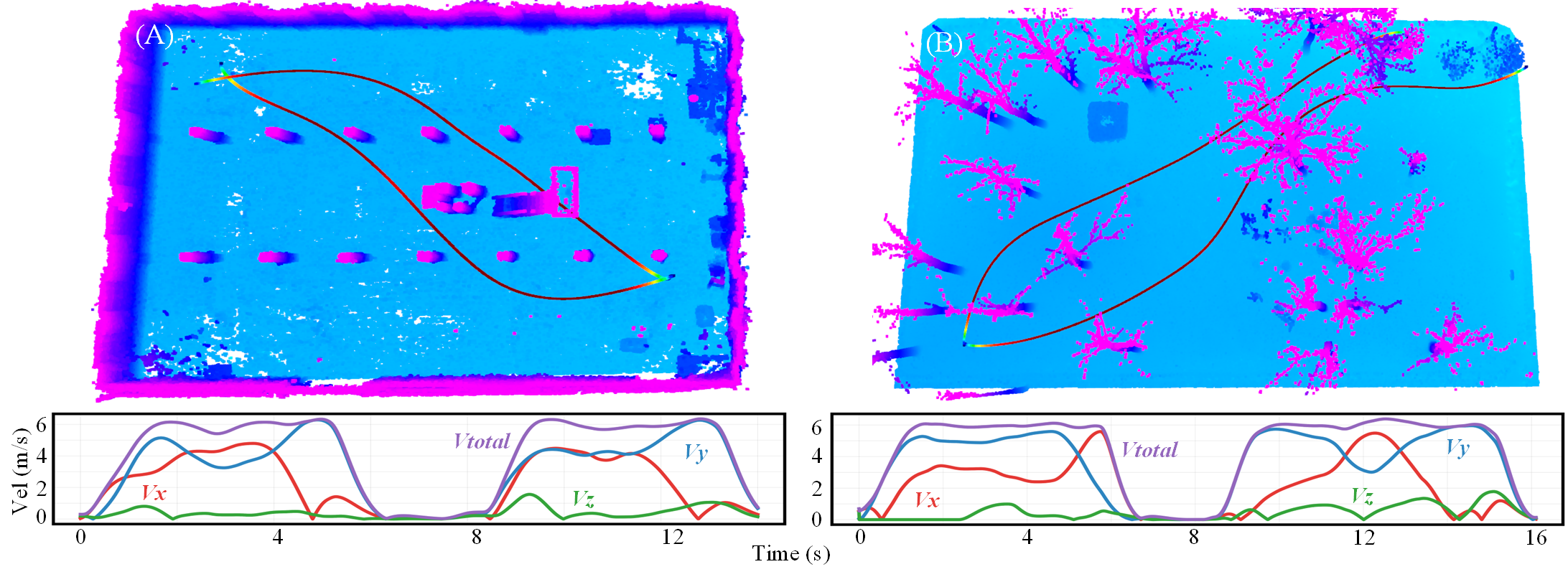}
    \vspace{-8mm}
    \caption{Simulated flight trajectories and velocity profiles of the proposed method in (A) the industrial facility and (B) the forest environment; the third-person views are shown in Fig.~\ref{fig:overview}(B) and Fig.~\ref{fig:overview}(A), respectively.}
    \label{fig:simexp}
    \vspace{-4mm}
\end{figure*}

\begin{figure*}[!t]
    \centering

    %================ Left: Figure ================
    \begin{minipage}[t]{0.48\textwidth}
        \vspace{0pt}
        \centering

        \includegraphics[width=\linewidth]{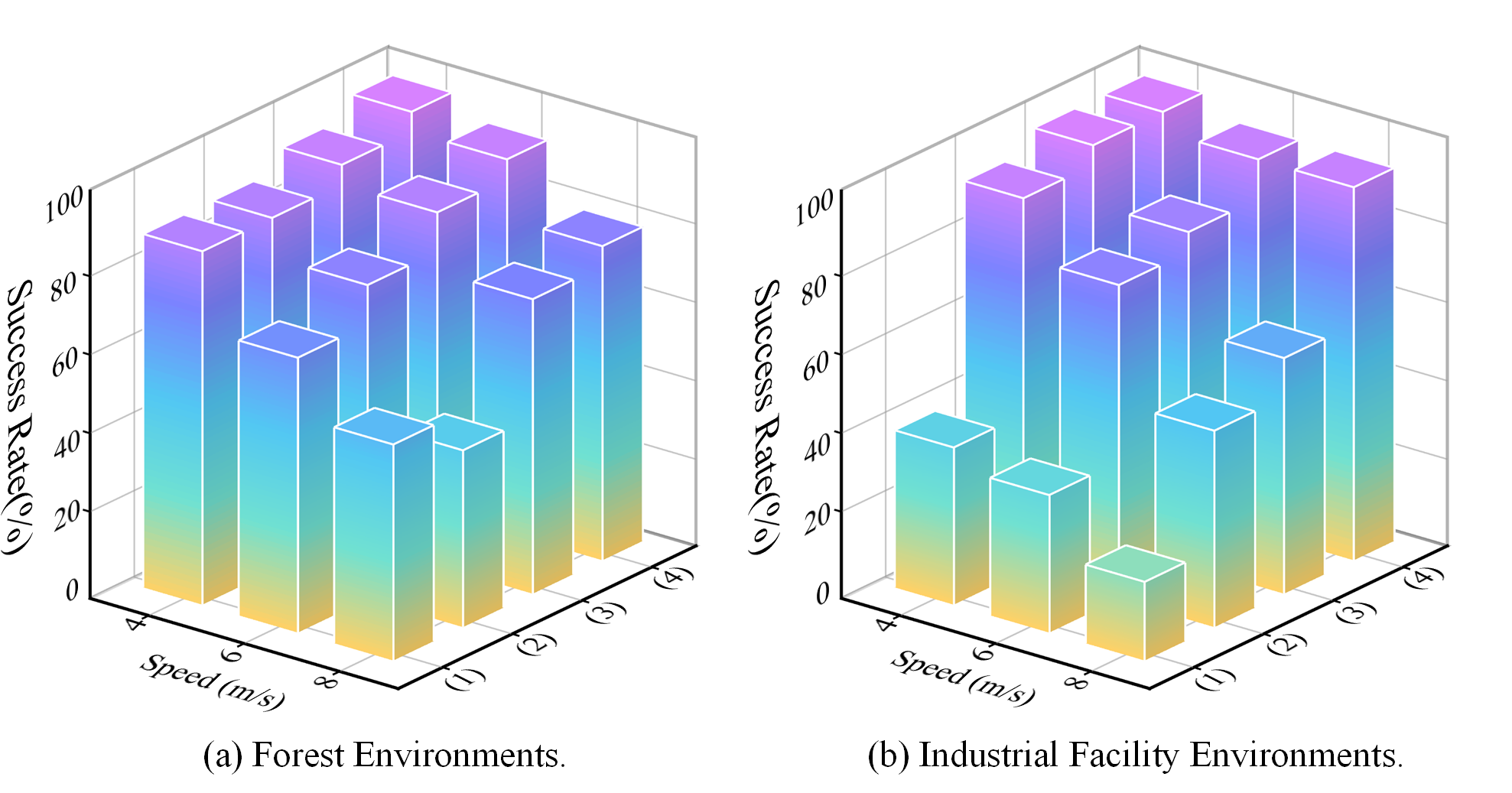}
        \vspace{-3mm}

        \captionof{figure}{Success rates under different desired speeds and environments. From front to back, the bars correspond to (1) RGB Baseline, (2) RGB + FA, (3) T-RGB + FA, and (4) Ours, respectively.}
        \label{fig:success}
    \end{minipage}%
    \hfill
    %================ Right: Table ================
    \begin{minipage}[t]{0.48\textwidth}
        \vspace{0pt}
        \centering

        \captionof{table}{Ablation Study.}
        \label{tab:ablation}

        \footnotesize
        \setlength{\tabcolsep}{2.5pt}
        \renewcommand{\arraystretch}{1.10}

        \resizebox{\linewidth}{!}{
        \begin{tabular}{lcccccc}
            \toprule
            \multirow{2}{*}{Method}
            & \multicolumn{3}{c}{Success Rate (\%) $\uparrow$}
            & \multicolumn{3}{c}{Average Speed (m/s) $\uparrow$} \\
            \cmidrule(lr){2-4}
            \cmidrule(lr){5-7}
            & Forest & Facility & Avg.
            & Forest & Facility & Avg. \\
            \midrule

            Depth Oracle$^\dagger$
            & 100.0 & 100.0 & 100.0
            & 5.18 & 4.12 & 4.65 \\
            \midrule

            RGB Baseline
            & 72.5 & 32.5 & 52.5
            & 3.95 & 5.18 & 4.56 \\

            RGB+FA
            & 80.0 & 70.0 & 75.0
            & 4.67 & 4.47 & 4.57 \\

            T-RGB+FA
            & 85.0 & 75.0 & 80.0
            & \textbf{5.30} & 5.12 & 5.21 \\

            \textbf{Proposed}
            & \textbf{90.0} & \textbf{92.5} & \textbf{91.3}
            & 5.29 & \textbf{5.30} & \textbf{5.30} \\

            \bottomrule
        \end{tabular}
        }

        \vspace{3mm}
        \parbox{\linewidth}{\footnotesize
        $^\dagger$Depth Oracle uses simulator-provided depth observations. Bold values indicate the best results among RGB-based planners. Average speed is computed over all trials.}
    \end{minipage}

    \vspace{-3mm}
\end{figure*}

\begin{figure}[t]
    \centering

    % --- Ablation figure ---
    \includegraphics[width=\linewidth]{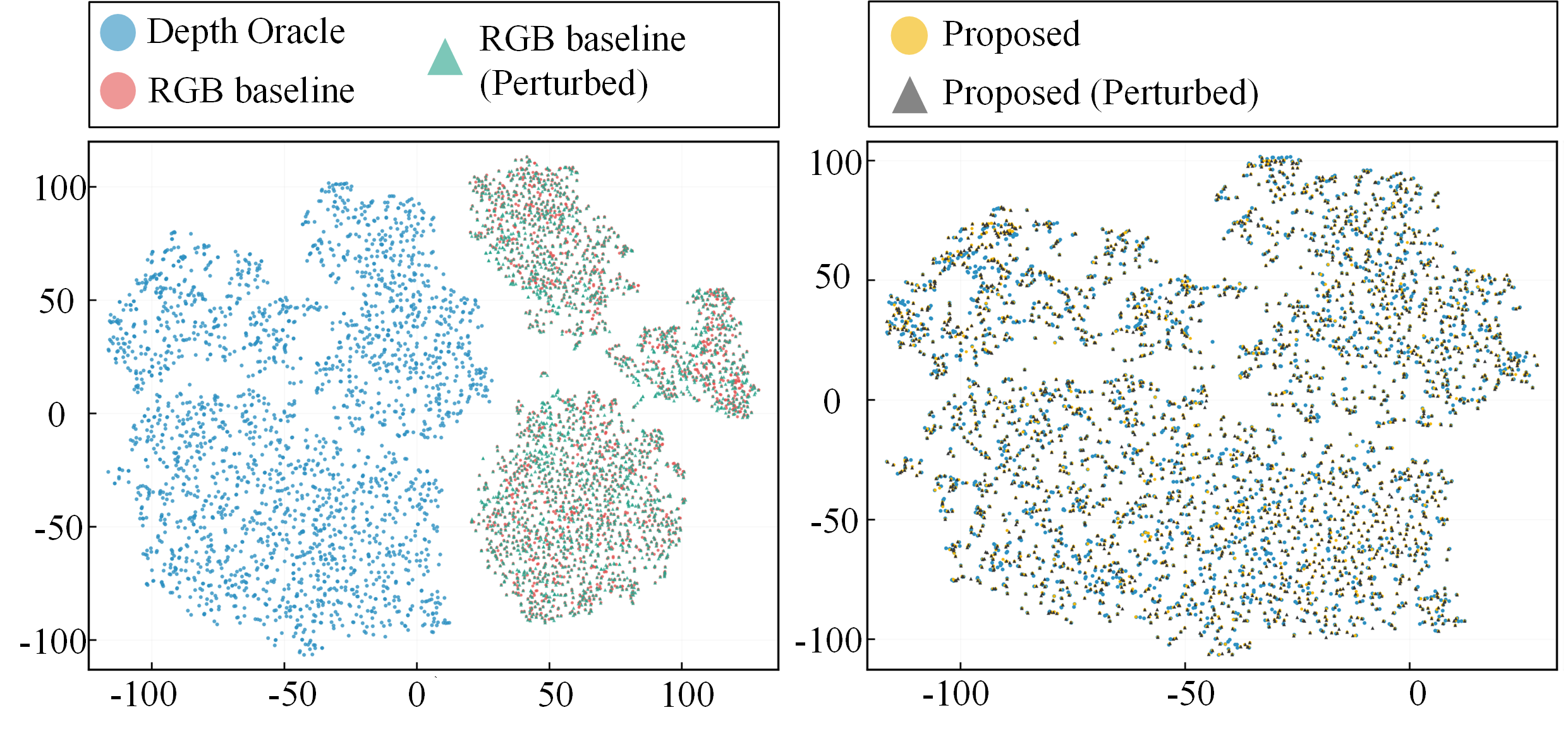}
    \vspace{-8mm}
    \caption{Visualization of planning-feature distributions using t-distributed stochastic neighbor embedding (t-SNE). All feature sets were concatenated, normalized using shared statistics, and jointly embedded in a single t-SNE run. Circles denote features extracted from nominal observations, whereas triangles denote features extracted from the same observations after evaluation-time appearance perturbations. For visual clarity, different subsets of the resulting shared embedding are displayed in two separate panels.}
    \label{fig:ablation}

    \vspace{8pt}

    % --- GSI table ---
    \footnotesize
    \setlength{\tabcolsep}{5pt}
    \renewcommand{\arraystretch}{1.12}

    \begin{threeparttable}
        \captionof{table}{GSI comparison of planning-feature distributions.}
        \label{tab:feature_gsi}

        \begin{tabular}{lcc}
            \toprule
            \multirow[c]{2}{*}{Method}
            & \multicolumn{2}{c}{GSI $\downarrow$} \\
            \cmidrule(lr){2-3}
            & Depth–RGB Separability $^\dagger$
            & Appearance Consistency $^\ddagger$ \\
            \midrule
            
            RGB Baseline
            & 1.0000
            & 0.1432 \\

            Proposed
            & \textbf{0.4134}
            & \textbf{0.0438} \\

            \bottomrule
        \end{tabular}

        \begin{tablenotes}[flushleft]
            \scriptsize
            \item[$\dagger$] GSI between the Depth Oracle and nominal RGB features.
            \item[$\ddagger$] GSI between the nominal RGB and appearance-perturbed RGB features.
        \end{tablenotes}
    \end{threeparttable}
    \vspace{-2mm}
\end{figure}

\subsubsection{Shared motion-primitive decoder}
Because all candidate primitives share the same body-frame state observation $\mathbf{x}$, each lattice primitive is assigned a learned token $\boldsymbol{\pi}_i$.  These tokens break the symmetry among otherwise identical state-conditioned queries and allow each decoder slot to specialize to its associated lattice direction.  Concatenating $\boldsymbol{\pi}_i$ with $\mathbf{x}$ and passing them through a shared MLP produces the primitive query $\mathbf{Q}_i$, where $i=1,\ldots,N$.  Stacking all primitive queries forms $\mathbf{Q}$. All queries are jointly processed by a Transformer decoder $\mathcal{T}_{P}$, whose self-attention captures relations among candidate primitives and whose cross-attention retrieves obstacle information from the position-encoded planning tokens:
\begin{equation}
    \mathbf{F}^{\mathrm{traj}}=\mathcal{T}_{P}\!\left(
    \mathbf{Q},\operatorname{PE}(\mathbf{F})\right),
    \qquad \mathbf{F}\in\{\mathbf{F}_D,\mathbf{F}_I\}.
    \label{eq:planning_decoder}
\end{equation}
A shared prediction head $\mathbf{MLP}_{\mathrm{out}}$ maps each decoded token to the normalized end state $\widehat{\mathcal{S}}_f^{(i)}$, selection score $s^{(i)}$, and predicted collision cost $\hat{c}^{(i)}$ of the corresponding primitive:
\begin{equation}
    \mathbf{MLP}_{\mathrm{out}}(\mathbf{F}_i^{\mathrm{traj}})
    =\left[\widehat{\mathcal{S}}_f^{(i)},
    s^{(i)},\hat{c}^{(i)}\right].
\end{equation}
Sharing the state encoder, planning decoder, and output head ensures that $\mathbf{F}_D$ and $\mathbf{F}_I$ support the same downstream planning computation, while aligning $\mathbf{F}_D$ and $\mathbf{F}_I$ transfers the expert's compact geometric representation.  This shared interface makes the feature alignment planning-oriented rather than purely appearance-oriented.

Finally, an auxiliary depth decoder $G_D$ maps the fused representation $\mathbf{F}_I$ to a coarse depth estimate,
$\widehat{D}_k=G_D(\mathbf{F}_I)$. This branch is used only during training to encourage $\mathbf{F}_I$ to preserve the geometry and spatial structure of the environment and is omitted during deployment. Consequently, online planning requires only the two RGB frames, their relative pose, and the MAV state.

\subsection{Progressive Three-Stage Training}
Training the proposed method involves two coupled objectives. The first is to learn a smooth and collision-aware trajectory planner. The second is to make the monocular branch learn the planning-relevant geometry available to the privileged depth expert. Jointly optimizing both objectives from scratch would require the network to learn geometry, temporal association, and trajectory planning simultaneously. We therefore formulate separate planning and cross-modal perception objectives and optimize them progressively in three stages.

\subsubsection{Differentiable planning objective}
For each candidate primitive $i$, the predicted end state and the current state uniquely determine a quintic polynomial trajectory.  We define the trajectory cost as:
\begin{equation}
    J^{(i)} = w_{j}J_{\mathrm{jerk}}^{(i)}
    +w_{c}J_{\mathrm{col}}^{(i)}
    +w_{g}J_{\mathrm{goal}}^{(i)} ,
    \label{eq:traj_cost}
\end{equation}
where the three terms evaluate motion smoothness, collision risk, and
goal progress, respectively.  

Following classical minimum-snap~\cite{r31} and polynomial trajectory optimization~\cite{r30}, we define the smoothness term as the integral of squared jerk along the polynomial trajectory:
\begin{equation}
    J_{\mathrm{jerk}}^{(i)}
    =\int_{0}^{T}\left\|
    \frac{\mathrm{d}^{3}\mathbf{p}^{(i)}(t)}{\mathrm{d}t^{3}}
    \right\|_{2}^{2}\,\mathrm{d}t,
    \label{eq:jerk_cost}
\end{equation}
where $\mathbf{p}^{(i)}(t)$ denotes the position along primitive $i$. Inspired by the ESDF-based obstacle potential in~\cite{r4}, the collision cost is evaluated at $M$ uniformly spaced time instants $t_m=mT/M$, $m=1,\ldots,M$:
\begin{equation}
\begin{split}
    J_{\mathrm{col}}^{(i)}
    &=\frac{1}{M}\sum_{m=1}^{M}
      \phi\!\left(d(\mathbf{p}^{(i)}(t_m))\right),\\
    \phi(d)&=\exp\!\left(-\frac{d-d_0}{r}\right),
    \label{eq:collision_cost}
\end{split}
\end{equation}
where $d(\cdot)$ is the ESDF distance to the nearest obstacle, $d_0$ is the safety margin, and $r$ controls the decay rate of the potential.

The goal cost encourages progress toward the local goal while penalizing lateral deviation:
\begin{equation}
\begin{aligned}
    J_{\mathrm{goal}}^{(i)}={}&\operatorname{SmoothL1}\!\left(\|\mathbf{g}-\mathbf{p}(0)\|_2,\mathbf{e}_g^{\top}\Delta\mathbf{p}^{(i)}\right)\\&+\eta_g\left\|(\mathbf{I}-\mathbf{e}_g\mathbf{e}_g^{\top})\Delta\mathbf{p}^{(i)}\right\|_2,
\end{aligned}
    \label{eq:goal_cost}
\end{equation}
where $\Delta\mathbf{p}^{(i)}=\mathbf{p}^{(i)}(T)-\mathbf{p}(0)$, $\mathbf{e}_g$ is the unit goal direction, and $\eta_g$ weights lateral deviation.  

Since the polynomial coefficients are uniquely determined by the current and predicted end states, all three costs are differentiable with respect to the predicted end state and can directly supervise trajectory prediction through back-propagation.  The trajectory loss $\mathcal{L}_{\mathrm{traj}}$ is the mean of $J^{(i)}$ over all candidate primitives and training samples.

In addition to optimizing the end states, we supervise the two scalar outputs associated with each primitive.  For the selection score, we mark the lowest-cost top-$K$ collision-free candidates as positive and apply a binary cross-entropy (BCE) loss to the predicted logits $\mathbf{s}$. For the collision cost, the network prediction is supervised using a
smooth $L_1$ loss:
\begin{equation}
    \mathcal{L}_{\mathrm{sel}} =\operatorname{BCE}(\sigma(\mathbf{s}),\mathbf{y}),
    \qquad
    \mathcal{L}_{\mathrm{safe}} =\operatorname{SmoothL1}
    (\hat{\mathbf{c}},{J}_{\mathrm{col}}).
    \label{eq:output_losses}
\end{equation}

Accordingly, at inference, the $K$ primitives with the highest selection scores $s^{(i)}$ are retained to form the candidate set $\mathcal{K}$. Each shortlisted primitive $i\in\mathcal{K}$ is then evaluated using $\hat{J}_{\mathrm{online}}^{(i)}=w_jJ_{\mathrm{jerk}}^{(i)}+w_c\hat{c}^{(i)}+w_gJ_{\mathrm{goal}}^{(i)}$. The final primitive is selected as $i^\star=\arg\min_{i\in\mathcal{K}}\hat{J}_{\mathrm{online}}^{(i)}$.

\subsubsection{Cross-modal geometric supervision}
The second objective transfers the planning representation of the frozen depth branch to the monocular RGB branch.  We align $\mathbf{F}_I$ with $\mathbf{F}_D$ using the feature-alignment loss $\mathcal{L}_{\mathrm{feat}}$:
\begin{equation}
\begin{split}
    \mathcal{L}_{\mathrm{feat}}
    ={}&\alpha_{\mathrm{cos}}
    \left[1-\cos\!\left(
    \mathbf{F}_I,
    \operatorname{sg}(\mathbf{F}_D\right)\right]\\
    &+\alpha_{\mathrm{abs}}\operatorname{SmoothL1}\!\left(
    \mathbf{F}_I,\operatorname{sg}(\mathbf{F}_D)\right),
    \label{eq:feature_distillation}
\end{split}
\end{equation}
where $\operatorname{sg}(\cdot)$ denotes stop-gradient.  The cosine term aligns the global feature direction, whereas the smooth $L_1$ term preserves the absolute feature values required by the shared planner.

Feature alignment constrains the compact planning representation but provides only indirect supervision of spatial geometry.  We therefore introduce auxiliary coarse-depth prediction as a complementary geometric regularizer for the fused temporal representation.  The objective combines a mean-normalized depth term that emphasizes relative scene structure with an absolute-depth term that penalizes depth-value errors.  Let $\mu(\cdot)$ denote the mean depth; the auxiliary objective is:
\begin{equation}
    \mathcal{L}_{\mathrm{depth}}
    =\beta_{\mathrm{rel}}
    \left\|
    \frac{\widehat{D}_k}{\mu(\widehat{D}_k)}
    -\frac{D_k}{\mu(D_k)}
    \right\|_1
    +\beta_{\mathrm{abs}}
    \left\|\widehat{D}_k-D_k\right\|_1.
    \label{eq:aux_depth_loss}
\end{equation}
The complete geometric supervision loss is
\begin{equation}
    \mathcal{L}_{\mathrm{geo}}=w_{f}\mathcal{L}_{\mathrm{feat}}+w_{d}\mathcal{L}_{\mathrm{depth}}.
    \label{eq:geometric_loss}
\end{equation}

\subsubsection{Three-stage optimization}
\textit{Stage I: depth-based planning pretraining.}
We first train the depth encoder together with the state encoder, shared motion-primitive decoder, and prediction head.  Single-frame depth observations provide explicit geometry, while trajectory costs provide planning supervision and the planning loss is $\mathcal{L}_{\mathrm{plan}}=\mathcal{L}_{\mathrm{traj}}+ \mathcal{L}_{\mathrm{sel}}+\mathcal{L}_{\mathrm{safe}}$. The monocular branch is not used, and the stage is optimized with $\mathcal{L}^{(1)}=\mathcal{L}_{\mathrm{plan}}$.

\textit{Stage II: cross-modal geometric supervision.}
We then freeze the depth branch and all shared planning modules.  Using paired depth observations and two-frame RGB images with their relative poses, we optimize the monocular branch solely with the geometric supervision
loss, $\mathcal{L}^{(2)}=\mathcal{L}_{\mathrm{geo}}$.

\textit{Stage III: RGB planning adaptation.}
Finally, privileged depth supervision is removed and the RGB perception branch is frozen.  Sequential visual observations and  trajectory costs are used to optimize the state encoder, shared motion-primitive decoder, and prediction head with
$\mathcal{L}^{(3)}=\mathcal{L}_{\mathrm{plan}}$.  This final stage adapts the downstream planner to the distilled RGB representation while preserving the geometry-aware visual features acquired in Stage II.

%\subsection{Simulation Environment}

\section{Experiments}

We evaluate the proposed method through both simulation and real-world experiments.  Our simulation framework consists of a photorealistic rendering module based on gsplat and a CUDA-accelerated geometric simulation module.  The former synthesizes monocular observations from reconstructed 3DGS scenes, whereas the latter performs ray casting against the corresponding environment point clouds to generate spatially aligned depth observations.  The constructed environments comprise autumn and summer forests and unfinished buildings. All test scenes are held out from training.

\subsection{Simulation Experiments and Ablation Study}

\begin{figure*}[!t]
    \centering

    %================ Left Figure ================
    \begin{minipage}[t]{0.48\textwidth}
        \vspace{0pt}
        \centering

        % Fixed-height image area
        \begin{minipage}[c][65mm][c]{\linewidth}
            \centering
            \includegraphics[
                width=\linewidth,
                height=65mm,
                keepaspectratio
            ]{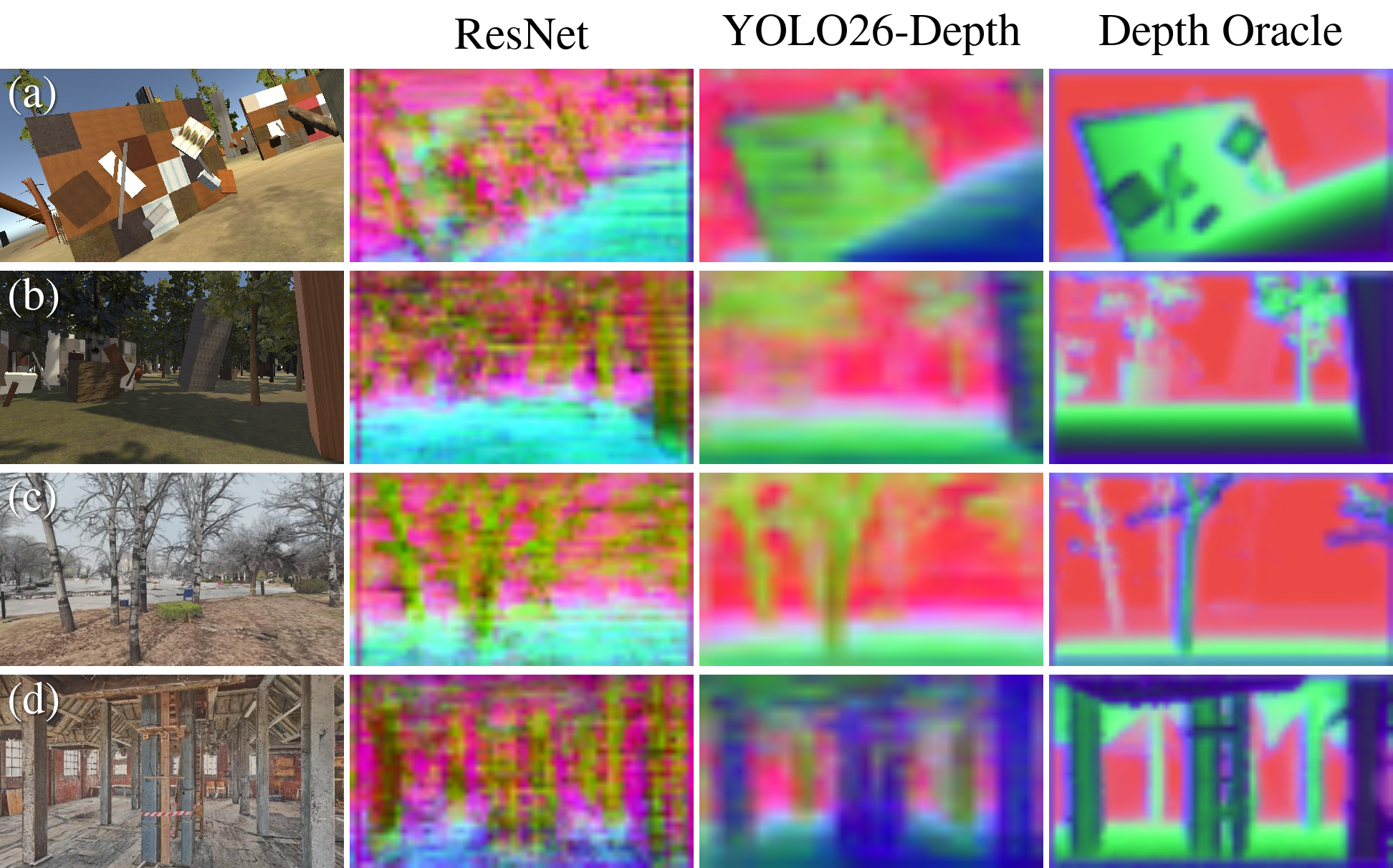}
        \end{minipage}
        \vspace{-6mm}
        \caption{Qualitative feature visualizations. Columns show the monocular image, temporally fused features from the ResNet and YOLO26-Depth implementations, and features from the Depth Oracle. Rows (a)--(b) show Unity scenes, while (c)--(d) show 3DGS scenes. Features are independently projected using principal component analysis (PCA).}
        \label{fig:yolofeature}
    \end{minipage}%
    \hfill
    %================ Right Figure ================
    \begin{minipage}[t]{0.48\textwidth}
        \vspace{5pt}
        \centering

        % Fixed-height image area
        \begin{minipage}[c][65mm][c]{\linewidth}
            \centering
            \includegraphics[
                width=\linewidth,
                height=65mm,
                keepaspectratio
            ]{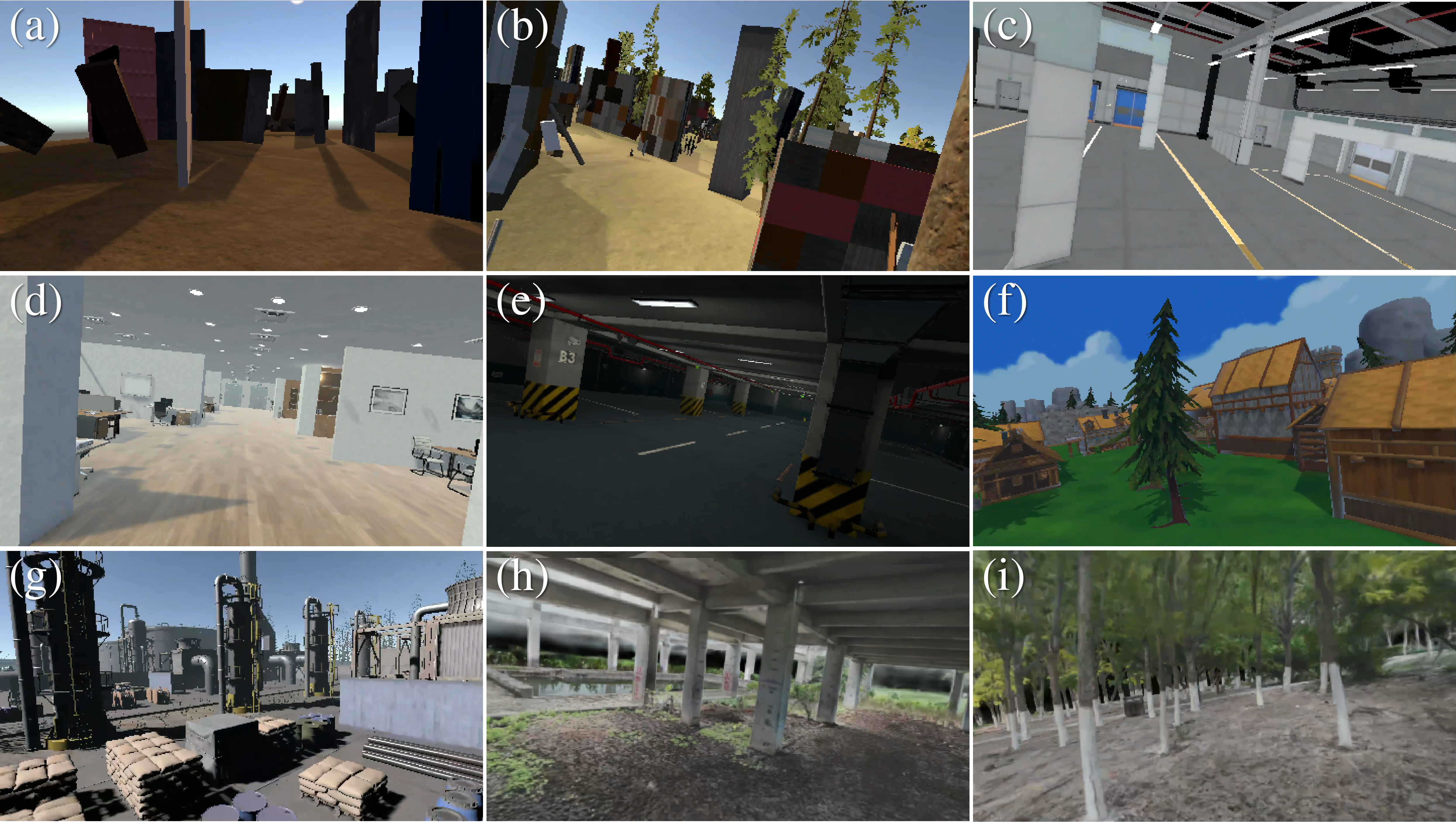}
        \end{minipage}
        \vspace{-7mm}
        \caption{Flight examples using YOLO26-Depth as the RGB encoder. Panels (a)--(b) show the two Unity training environments. Panels (c)--(g) show five Unity test environments, and panels (h)--(i) show two 3DGS test environments.}
        \label{fig:unityand3dgs}
    \end{minipage}
    \vspace{-3mm}
\end{figure*}

To evaluate the flight performance of our method, we conduct experiments in two representative environments: 1) a cluttered, unstructured forest and 2) a structured industrial facility. We evaluate four variants, whose configurations are detailed below. For each environment and variant, we conduct 20 trials at each desired flight speed of $4$, $6$, and $8\,\mathrm{m/s}$ and report the corresponding success rates. As shown in Figs.~\ref{fig:overview}(A)--(B), ~\ref{fig:simexp}, and~\ref{fig:success}, the proposed method achieves fast and reliable flight in both forest and industrial facility environments.

To further examine the contributions of temporal pose input, feature alignment supervision and the auxiliary coarse-depth prediction, we evaluate the following five variants: 
1)~\textbf{Depth Oracle}, which plans from simulator-provided depth observations and serves as a reference for planning with privileged geometric information;
2)~\textbf{RGB Baseline}, which receives a single monocular image and is trained end-to-end using only the planning objective;
3)~\textbf{RGB + Feature Alignment}, which additionally aligns its planning features with those produced by the privileged depth branch;
4)~\textbf{Temporal RGB + Feature Alignment}, which uses the same network architecture as the proposed method but excludes the auxiliary depth supervision in Stage II;
5)~\textbf{Proposed}, which takes two consecutive monocular images and their relative pose as input and is trained using both planning-alignment supervision and an auxiliary coarse-depth prediction objective. All variants are evaluated in the industrial facility and cluttered forest at a desired speed of $6\,\mathrm{m/s}$. Each model is evaluated over 40 trials in each environment. 
% As summarized in Table~\ref{tab:ablation}, PlanBridge achieves the best combination of safety and flight speed among the RGB-based planners. Comparing Baseline with RGB+FA shows that supervision from the privileged depth branch consistently improves planning performance. This result indicates that the privileged planning representation provides the RGB branch with geometric cues that are more relevant to collision avoidance.
% Furthermore, PlanBridge outperforms T-RGB+FA in both success rate and flight speed, supporting the effectiveness of the auxiliary depth reconstruction objective.  Unlike alignment performed only in the compact planning-feature space, this auxiliary objective directly constrains the recovery of spatial structure, thereby encouraging the network to exploit cross-frame motion and relative pose when extracting planning-relevant geometric information.
Average speed is computed over all trials, including unsuccessful ones. As summarized in Table~\ref{tab:ablation}, RGB+FA increases the average success rate from 52.5\% to 75.0\% relative to RGB Baseline, while producing a nearly unchanged average speed. This result indicates that planning-feature alignment primarily improves navigation reliability. Adding the pose-conditioned temporal input, as represented by T-RGB+FA, further increases the success rate to 80.0\% and the average speed to $5.21\,\mathrm{m/s}$. Finally, auxiliary depth supervision raises the success rate to 91.3\% while maintaining a comparable average speed of $5.30\,\mathrm{m/s}$. These controlled comparisons suggest that feature alignment, pose-conditioned temporal fusion, and auxiliary geometric supervision provide complementary benefits.

\begin{figure*}[!t]
    \centering
    \includegraphics[width=\textwidth]{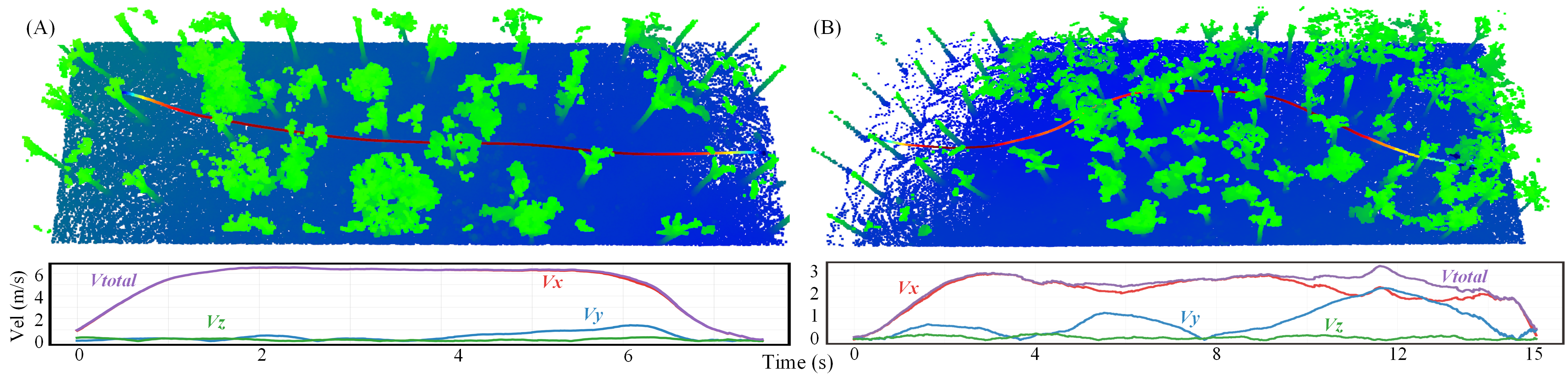}
    \vspace{-4mm}
    \caption{Trajectory visualizations of the real-world experiments. (A) Obstacle-avoidance flight, reaching a maximum speed of $6.5\,\mathrm{m/s}$. (B) Agile target-tracking flight. The third-person views are shown in Fig.~\ref{fig:overview}(C) and Fig.~\ref{fig:overview}(D), respectively.}
    \label{fig:realexp}
    \vspace{-2mm}
\end{figure*}

We additionally visualize the latent planning features produced by Depth Oracle, RGB Baseline, and the proposed method using t-SNE. As shown in Fig.~\ref{fig:ablation},  the features produced by the proposed method exhibit greater distributional consistency with those of the privileged depth branch, whereas the directly trained baseline remains strongly modality-specific. This consistency is further quantified using the geometric separability index (GSI), which measures the separability between the RGB planning features and the privileged depth features in the original feature space. This observation suggests that, by guiding the RGB branch toward the planning representation learned by the depth branch, the proposed feature-bridging strategy encourages it to encode more planning-relevant geometric cues relevant to collision avoidance.

Beyond cross-modal alignment, we further examine the robustness of the learned representations to appearance variations by evaluating the proposed method under controlled appearance perturbations, including changes in exposure, color, contrast, and simulated sensor noise. The triangular markers in Fig.~\ref{fig:ablation} denote features extracted from these perturbed observations. Specifically, we apply perturbations stronger than those used for training augmentation to the same visual observations and compare the planning-feature distributions of Baseline and the proposed method before and after perturbation.  We regard the nominal and perturbed samples as two appearance conditions and quantify their separability using GSI.  A lower GSI indicates that the nominal and perturbed features are more difficult to distinguish. GSI is computed directly in the original feature space, whereas t-SNE is used only for visualization. As shown in Table~\ref{tab:feature_gsi}, our method achieves a lower GSI between nominal and perturbed features than RGB Baseline, indicating that it yields more consistent planning representations under appearance perturbations. 

Building on these results, we further explore transferring visual representations learned from large-scale datasets to the RGB planner. Diversity in scene structure and visual appearance is important for generalization, yet acquiring sufficiently diverse navigation data through our own data collection is costly. Given the depth-oriented pretraining and low inference latency of YOLO26-Depth~\cite{yolo26}, we adopt an RGB encoder that leverages its pretrained weights, replacing the ResNet trained on our limited dataset. To adapt these representations to navigation, we insert lightweight residual adapters after selected feature blocks in the backbone and neck and train them on our dataset using the proposed cross-modal geometric supervision. The adapted features are then passed to the existing pose-conditioned temporal fusion and planning modules.

Although the preceding analyses show that the proposed ResNet-based RGB planner achieves greater distributional consistency with the Depth Oracle than the RGB Baseline at the shared planning interface, such consistency does not necessarily imply clearly discernible scene structure in intermediate feature visualizations. Fig.~\ref{fig:yolofeature} compares the temporally fused features from the ResNet and YOLO26-Depth implementations with those of the Depth Oracle. The Depth Oracle features exhibit recognizable obstacle boundaries and spatial partitions corresponding to the scene layout. Similar structural patterns are visible in the pretrained RGB features across the illustrated Unity and 3DGS scenes, whereas obstacle structures are less discernible in the ResNet features. These observations suggest that the YOLO26-Depth implementation captures obstacle-related structural cues that visually resemble those of the Depth Oracle across scenes with different appearances.

To complement the feature-level analysis with closed-loop flight examples, we conduct an additional evaluation across environments. For this evaluation, we train the planner using data collected from two cluttered Unity environments and then evaluate it in five other Unity environments and two 3DGS environments. The Unity tests examine applicability across different scene layouts and visual appearances, while the 3DGS tests additionally introduce a change in rendering domain. Fig.~\ref{fig:unityand3dgs} presents the training environments and representative frames from flight recordings in the test environments. The planner trained in the two Unity environments enabled the MAV to fly and execute obstacle-avoidance maneuvers in the other seven environments.

\subsection{Real-World Flight}

Finally, we validate the proposed ResNet-based RGB planner through direct sim-to-real deployment. All samples used to train the planner for real-world deployment are collected from simulated rollouts in 3DGS-reconstructed environments, comprising forest sites spanning autumn and summer appearances and unfinished-building sites. Neither of the real-world flight sites is included in the training set. Furthermore, no data captured by the test platform's onboard camera or IMU are used for network training or fine-tuning. 

Our physical platform is a compact quadrotor measuring 119\,mm diagonally and weighing 750\,g. It is equipped with an NVIDIA Jetson Orin NX and a monocular fisheye camera that captures 1080p images. For planning, the fisheye observations are undistorted and reprojected into cylindrical images with a resolution of $640\times240$ and a $160^\circ\times55^\circ$ FOV, while VINS-Mono uses the original calibrated fisheye stream for state estimation. After TensorRT deployment, the planning network achieves an inference latency of $9.5\,\mathrm{ms}$, and the planner operates at 15\,Hz. The complete state estimation, planning, and control pipeline runs onboard the vehicle.

We first evaluate high-speed obstacle avoidance along an approximately 40\,m route through a dense forest. A goal point is specified before flight, after which the vehicle autonomously generates collision-aware trajectories and navigates toward the goal without human intervention. The desired speed is progressively increased from $3$ to $5$ and $6\,\mathrm{m/s}$ across repeated flights. At a desired speed of $6\,\mathrm{m/s}$, the vehicle reaches a maximum speed of $6.5\,\mathrm{m/s}$. As shown in Fig.~\ref{fig:realexp}(A), the vehicle rapidly accelerates to the desired speed and maintains high-speed flight through the cluttered route.

We further examine the replanning agility of the proposed method under a continuously changing goal direction. A downstream YOLO-based person detector updates the target direction online, and the vehicle is commanded to remain oriented toward the detected person while flying at a desired speed of $3\,\mathrm{m/s}$. Rather than evaluating the accuracy of the person-tracking module itself, this experiment serves as a test of the planner under frequent goal updates. As shown in Fig.~\ref{fig:realexp}(B), the changing target direction produces substantial variations among the velocity components, requiring the planner to continuously generate new trajectories while maintaining stable flight.

\section{Conclusions}
This paper presented a monocular visual planning framework for agile MAV flight. The proposed method uses the planning representation learned by a privileged depth branch as cross-modal supervision, extracts geometric cues by fusing two consecutive images with their relative pose, and further constrains the temporal representation through auxiliary coarse-depth prediction. Experiments demonstrate that the proposed method reduces discrepancies between monocular and depth-based planning features and feature-distribution shifts under appearance perturbations. Real-world experiments further demonstrate that the planner transfers directly from 3DGS-based simulation to real-world environments without hardware-specific data collection or real-world fine-tuning, supporting autonomous obstacle avoidance and agile person tracking.
\bibliographystyle{ieeetr}
\bibliography{ref}
\end{document}